\documentclass{bmvc2k}

\usepackage{graphicx}
\usepackage{tikz}
\usepackage{comment}
\usepackage{amsmath,amssymb} 
\usepackage{color}
\usepackage{booktabs}
\usepackage{algorithm}
\usepackage{algpseudocode}
\usepackage{multicol}
\usepackage{microtype}

\title{Darker than Black-Box: Face Reconstruction from Similarity Queries} %
\addauthor{Anton Razzhigaev}{anton.razzhigaev@skoltech.ru}{1}
\addauthor{Klim Kireev}{klim.kireev@epfl.ch}{1}
\addauthor{Igor Udovichenko}{igudaff@gmail.com}{2}
\addauthor{Aleksandr Petiushko}{petyushko@yandex.ru}{2} 

\addinstitution{
 Skolkovo Institute of Science and Technology\\
 Moscow, Russia
}
\addinstitution{
 Lomonosov Moscow State University\\
 Moscow, Russia
}

\runninghead{Razzhigaev, Kireev, Udovichenko, Petiushko}{Darker than Black-Box}

\begin{document}

\maketitle

\begin{abstract}
Several methods for inversion of face recognition models were recently presented, attempting to reconstruct a face from deep templates. Although some of these approaches work in a black-box setup using only face embeddings, usually, on the end-user side, only similarity scores are provided. Therefore, these algorithms are inapplicable in such scenarios. We propose a novel approach that allows reconstructing the face querying only similarity scores of the black-box model. While our algorithm operates in a more general setup,  experiments show that it is query efficient and outperforms the existing methods.

\end{abstract}

\begin{figure}[t]
\begin{center}
   \includegraphics[width=0.75\linewidth]{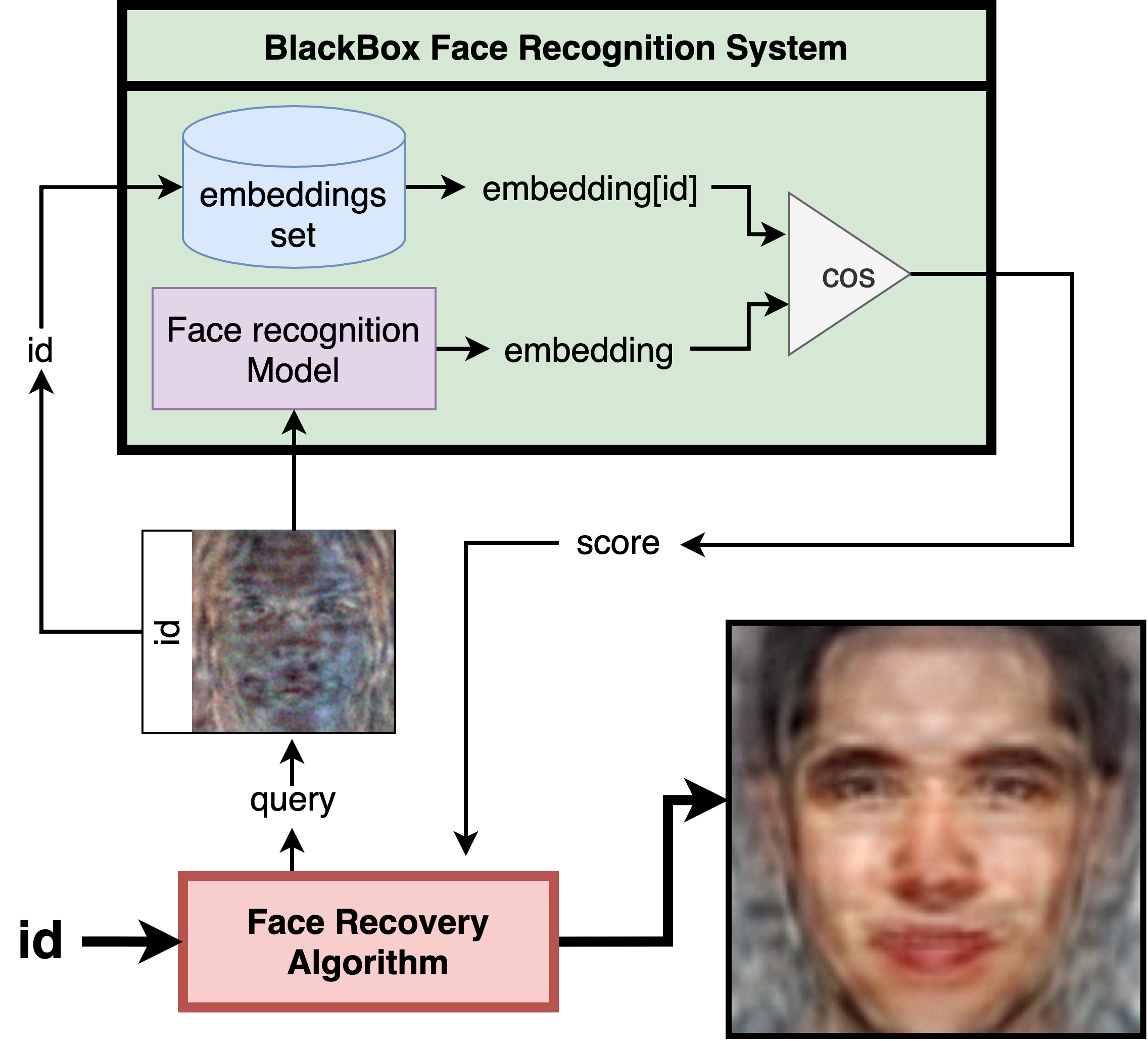}
\end{center}
    \caption{The schematic of the face recovery from the similarity queries to the face recognition system. Iteratively requesting a similarity score between a target person id with a specially generated set of images it is possible to reconstruct the appearance of this person.}
\label{fig1}
\end{figure}

\section{Introduction}


Deep neural networks (DNNs) are now the leading instrument in artificial intelligence research.
DNNs have achieved state-of-the-art performance in many machine learning areas including computer vision, natural language processing, etc.

Face recognition has attracted large attention over the last years~\cite{Wang_2021}.
Face recognition systems are widely used now in various applications such as security systems and access control.
Also, modern data protection legislation pays special attention to data leakage problems.
Consequently, there are special requirements for the security of face recognition systems.
The insufficient reliability of such systems is a significant source of risk. 
Thus, it is crucial to study the robustness and privacy issues of the models that are used in critical applications.

Modern face recognition pipelines consist of several steps~\cite{Li2011handbook}.
First, the face is detected in the given image, and the alignment process is done.
Then, the aligned face is fed to the face identification network, which encodes the aligned face to the embedding~--- the feature vector of lower dimensionality.
Last, this vector can be classified or compared with the embedding of the target face.

One of the best and most popular publicly available models for the identification step of face recognition is ArcFace~\cite{deng2019arcface}.
It uses ResNet~\cite{he2015deep} backbone and additive angular margin loss to improve the discriminate power of the face recognition model and to stabilize the training process.

Face recognition systems might be attacked from various perspectives~\cite{liu2016maskface,patel2016secureface,sharif2016faceattacks}.
This paper focuses on data leakage vulnerability where a face embedding can be reconstructed to a recognizable face.

In our work, we consider the novel fully black-box problem statement, where neither any information about the face identification model nor deep features of face images is used, but only the cosine similarity between images produced by a black-box model. Although the proposed algorithm can work with embedding-based queries and in a fully white-box setup. We present the face recovery algorithm that works in such a restrictive scenario and achieves competitive performance in the previously considered problem statements. We are inspired by the concept of ``eigenface'' first proposed in works~\cite{kirby1990klp,turk1991eigenfaces} for the identification problem.
Eigenfaces can be defined as principal components of the set of face images or some other set of face-like images that explain most data variations.

The scheme of the face reconstruction pipeline is described in Fig. \ref{fig1}. Attacker queries the face recognition system with the generated images to acquire the similarity score with some certain id. After that, it generates the face image which posses the similarity value and allows to identify the underlying person with this id. 

\textbf{Main contributions of this work are the following:}
\begin{itemize}
    \item The novel face reconstruction method was proposed operating with only similarity value queries;
    \item We achieve state-of-the-art performance on several face reconstruction benchmarks;
    \item The method is query efficient and conceptually simple;
    \item The method achieves not only visual similarity but also retains a high score concerning the independent face recognition system;
    \item In comparison to closely related
    method~\cite{razzhigaev2020blackbox} an ability to recover color faces was added.
\end{itemize}

The source code is available on the Internet\footnote{\url{https://github.com/papermsucode/fullbboxrestoration}}.

\begin{table}
\caption{Comparison table}
\label{comp_table}
\bigskip
\centering
\begin{tabular}{lcccc}
\toprule
Algorithm & Target model & Setting &  Color & Input \\
\midrule
Ours & Arcface & Black-Box & + & Cos \\ 
Gaussian blobs~\cite{razzhigaev2020blackbox} & Arcface & Black-Box & - & Cos, norm\\ 
NBNet\cite{Mai_2019} & FaceNet & Black-Box & + & Embedding \\
Gradient wrt input \cite{mahendran2015understanding} & Any classifier & White-box & + & Whole model\\ 
\bottomrule
\end{tabular}
\end{table}

\section{Related Work}

Many types of attacks on face recognition systems and DNNs are known at the moment~\cite{akhtar2018advattacks}.
The first approach to restore a face from its embedding is white-box gradient back-propagation with respect to the input. 
This method is used in many works on the neural networks interpretability and the general class inversion~\cite{erhan2009visualizing,simonyan2013deep}, where the
knowledge of the model's architecture and the ability to run a backward pass is required.
In addition, due to the high dimensionality of images, it is necessary to add image priors, such as Total Variation~\cite{mahendran2015understanding} or Gaussian Blur~\cite{yosinski2015understanding}.
An application of this approach to the face restoration problem can be found in work~\cite{fredrikson2015faceinversion}. However, most works consider this type of model inversion only for closed-set problems, such as classification with the predefined number of classes. 

Another approach is the training-based inversion~\cite{dosovitskiy2016generating,dosovitskiy2016inverting,nash2019inverting}, where the
additional neural network is used to map face embedding to images.
The whole system is trained in the encoder-decoder style with $L_1$ or  $L_2$ loss between the original and reconstructed image.
The reconstruction process requires one forward pass through the decoder.
The original face recognition model is used as an encoder part and the additional network plays the role of a decoder,
therefore the back-propagation through the encoder part is also required in this approach.

The third category of face reconstruction methods is the black-box one that requires neither backward passes through the attacked model, nor any knowledge about its architecture.
In~\cite{Mai_2019} it was proposed to use a neighborly de-convolutional neural network (called NBNet) to reconstruct the recognizable face.
The training process is the following: the cropped and aligned face is fed to the black-box feature extractor.
Then NbNet is trained to map these features to images, and a perceptual loss~\cite{johnson2016perceptual} is used as a measure of distance between the original and reconstructed images.
Another black-box face restoration approach was explored in~\cite{razzhigaev2020blackbox}.
The algorithm is zeroth-order optimization in the space of 2D Gaussian functions.
A face image is assumed to lie in the class of clipped linear combinations of Gaussian blobs.
In that work, for the given embedding the restored face is the one that minimizes the weighted sum of cosine similarity between the target embedding and the output of the face feature extractor and $L_2$ loss between norms of embeddings as a regularizer.
A zeroth-order gradient projection method is used to find the solution, where the directions for the gradient estimation are sampled from the class of Gaussian blobs. 
In~\cite{duong2020vec2face} there was introduced a distillation framework (DiBiGAN) with learning the bijective mapping between images and their latent representations, and both black-box and white-box settings were considered.

The short comparison of the methods can be seen in Table \ref{comp_table}.

\section{Face recovery with eigenfaces sampling algorithm}
In this section, we introduce the algorithm for revealing faces from just face recognition similarity queries. We are inspired by Gaussian sampling algorithm \cite{razzhigaev2020blackbox} and eigenfaces concept \cite{kirby1990klp}, \cite{turk1991eigenfaces}. We use a set of eigenfaces generated in advance from publicly available data which does not intersect with the testing data and perform a zeroth-order optimization in this space.

\subsection{Eigenfaces generation}

\begin{figure}[t]
\begin{center}
   \includegraphics[width=0.9\linewidth]{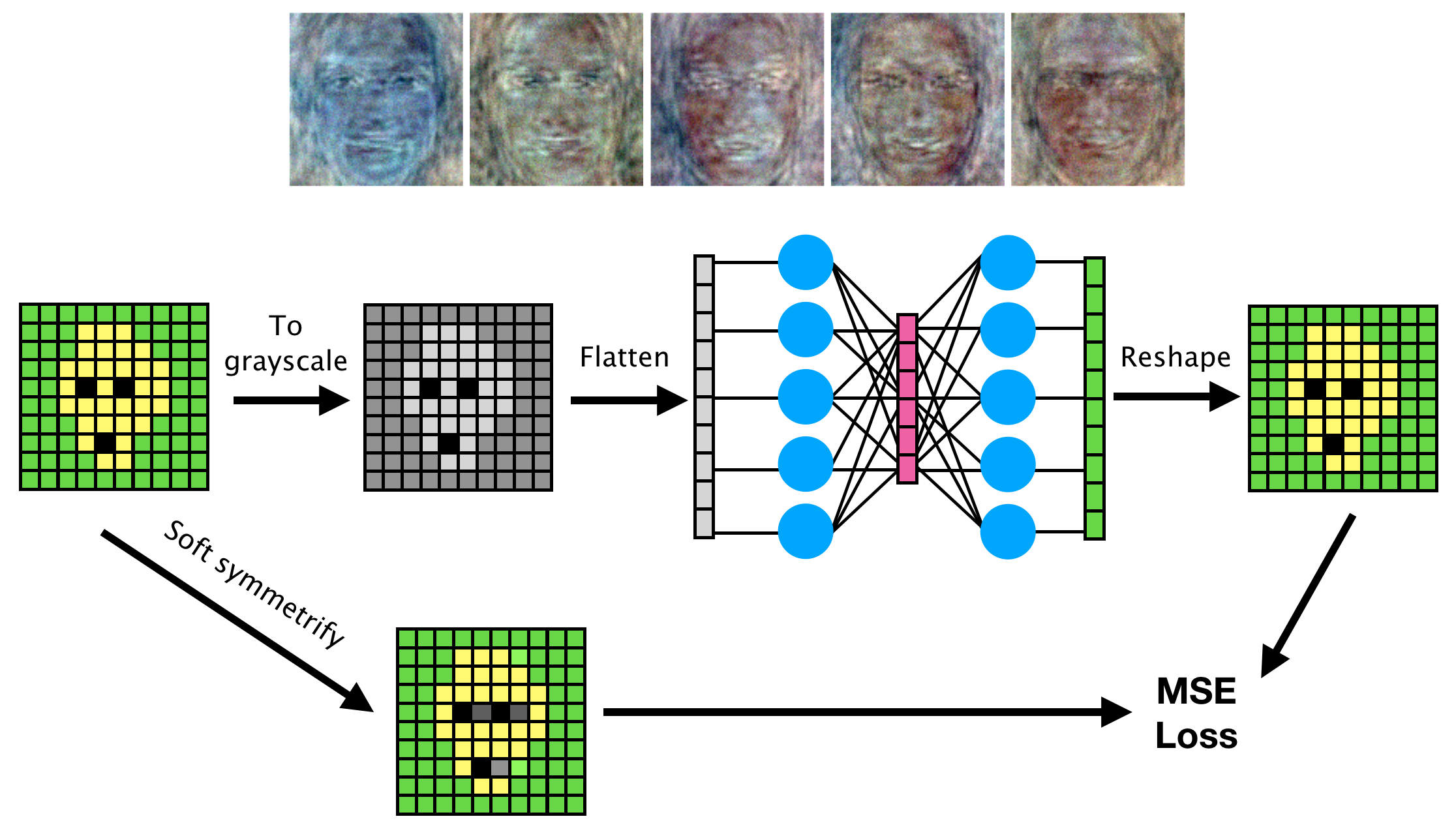}
\end{center}
   \caption{A linear autoencoder training scheme (without generative regularizing term for convenience). The eigenfaces are the reshaped parameters of the decoder part.}
\label{ae}
\end{figure}

The set of eigenfaces serves as the approximate basis in the space of faces (in fact, a linear hyperplane) and allows efficient optimization in this space retaining a very strong prior. In Fig. \ref{ae} some random eigenfaces obtained by the procedure described below are shown. We obtain a set of 1024 eigenfaces by training a linear autoencoder on the CelebA dataset. Actually, the eigenfaces are weights of a fully connected layer from the decoder part of the autoencoder.

The encoder end decoder consists of one linear layer with turned-off biases.

 \begin{equation}
 Y = W_{2}\cdot (W_{1}^{T} \cdot X),
 \end{equation}
 where,\\
\hspace*{3em}
\begin{tabular}{ll}
    $X$ --- the input flattened image with shape $(3 \cdot dim_y \cdot dim_x, 1)$,\\
    $W_1$ --- encoder weights with shape $(3 \cdot dim_y \cdot dim_x,1024)$,\\
    $W_2$ --- decoder weights with shape $(3 \cdot dim_y \cdot dim_x,1024)$,\\
    $Y$ --- the output flattened image with shape $(3 \cdot dim_y \cdot dim_x, 1)$,\\
    $dim_x,dim_y$ --- the horizontal and vertical sizes of the image.\\
\end{tabular}
\hspace*{3em}

The loss function used for training the autoencoder is partially shown in Fig. \ref{ae} and more formally can be described by:
 \begin{equation}
 L(X) =  \text{MSE}\left(\frac{X+2\text{R}(X)}{3}, W_{2}\cdot (W_{1}^{T} \cdot X)\right) + \text{MSE}(X,W_2 \cdot z ),
 \end{equation}
 where\\
\hspace*{3em}
\begin{tabular}{ll}
    \text{MSE} --- the mean squared error,\\
    R --- a vertical reflection operator,\\
    $z$ --- a random 1024-d vector sampled from $N(0,1)$ 
    
\end{tabular}
\hspace*{3em}\\

\subsection{Soft symmetry constraints}
As it was shown in \cite{razzhigaev2020blackbox} the symmetric constraints on generated faces demonstrate better recovery results. In our work, we use the soft symmetry constraint which is applied during the training of the autoencoder ($X\rightarrow \frac{X+2\text{R}(X)}{3}$) to make eigenfaces more symmetric. The effect of symmetrization can be seen in Fig. \ref{loss}.

\subsection{Generative regularisation}
\begin{figure}[t]
\begin{multicols}{2}
    \hfill
    \includegraphics[width=0.95\linewidth]{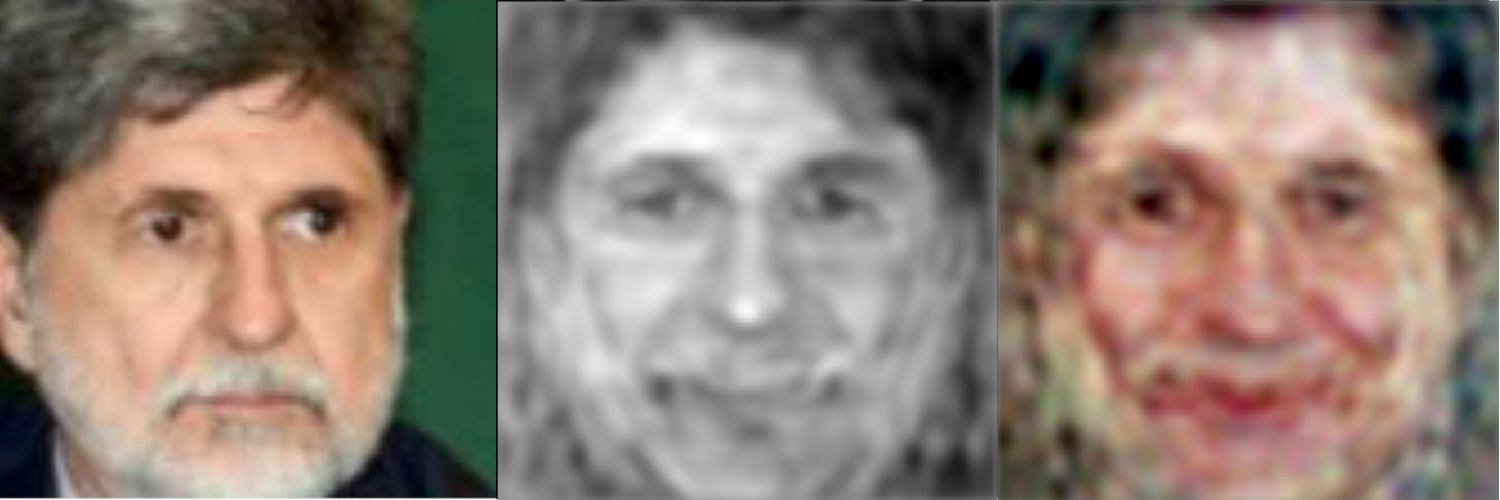}
    \hfill
    \caption{From left to right: original face, recovered from gray-scale eigenfaces, recovered from colorful eigenfaces.}
    \label{color_gray}
    \hfill
    \includegraphics[width=0.95\linewidth]{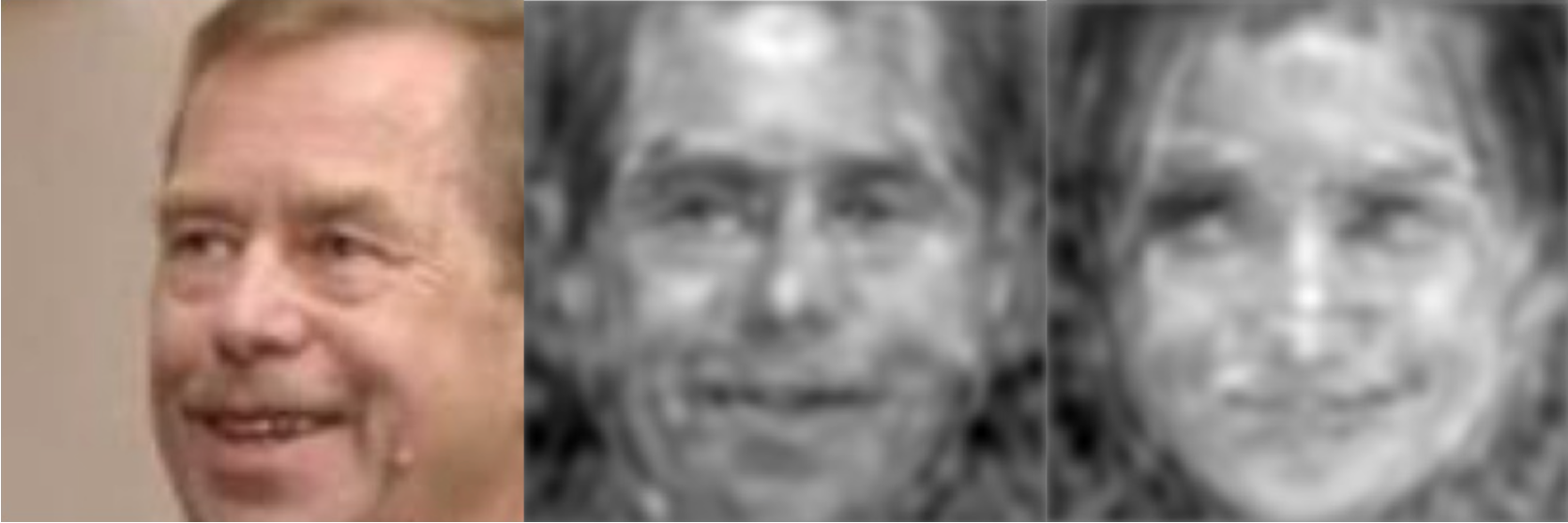}
    \hfill
    \caption{From left to right: original face, recovered with 10 restarts, recovered without restarts (got stuck in the local minimum -- wrong identity).}
\label{compare_multi}
\end{multicols}
\begin{center}
   \includegraphics[width=0.95\linewidth]{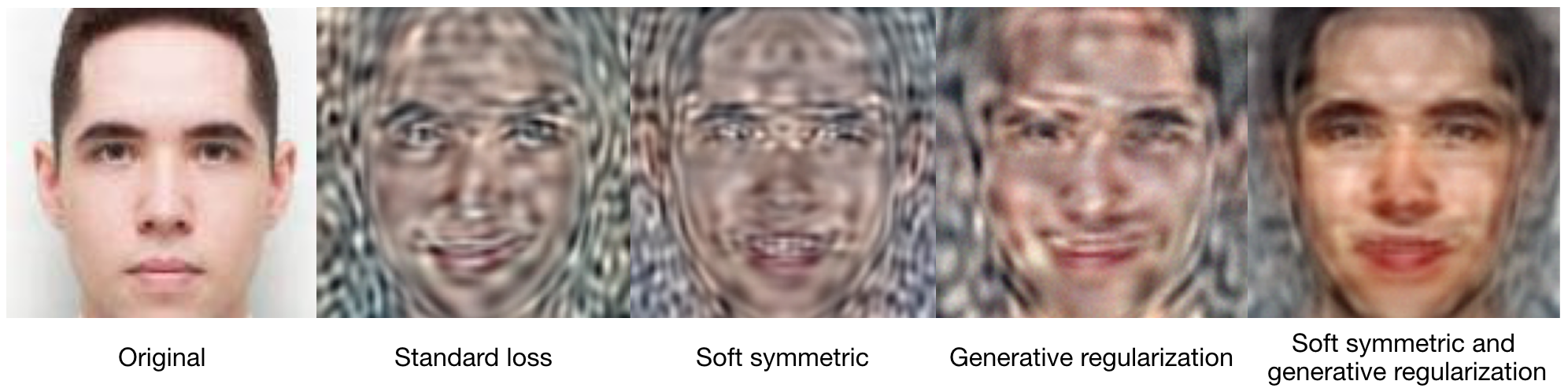}
\end{center}
   \caption{Face recovery from eigenfaces obtained from autoencoder with different types of losses.}
\label{loss}
\end{figure}

During the training of the eigenfaces set we use the special generative regularization term in the loss. We enforce the decoder to generate face-like images from any Gaussian noise using mean square loss between a random face $X$ and a decoded Gaussian noise: $\text{MSE}(X, W_2 \cdot z )$. The effect of different loss types can be seen in Fig. \ref{loss}.

\subsection{Colorful reconstruction}
Even though face embeddings do not contain much information about color \cite{razzhigaev2020blackbox}, eigenfaces have a specific natural color bias that can help to reconstruct colorful faces. We conducted several experiments with colorful and gray-scale recovery, demonstrating that our algorithm can reconstruct colorful faces. The comparison of gray-scale and colorful recovery is shown in Fig. \ref{color_gray}.

\subsection{Algorithm}

\begin{algorithm}[H]
\caption{Face recovery algorithm}\label{alg}
\textbf{INPUT:} target $id$, black-box system $S$, $N_{queries}$, eigenfaces matrix $\textbf{E}$
\begin{algorithmic}[1]
\State $X\leftarrow 0$
\For{$i\leftarrow 0$ to $N_{queries}$}:
\State Allocate image batch \textbf{X}
\State Sample batch $\textbf{Z}$  of random vectors
\State $\textbf{X}_j = X + \textbf{E} \cdot \textbf{Z}_j$
\State $\textbf{s}' = S(\textbf{X},id)$
\State $\text{ind} = \text{argmax}(\textbf{s}')$
\State $X \leftarrow X + \textbf{E} \cdot \textbf{Z}_{ind}$
\State  $i \leftarrow i + \text{batchsize}$
\EndFor
\end{algorithmic}
\label{algorithm}
\textbf{OUTPUT:} reconstructed face $X$
\end{algorithm}

The recovery procedure is a zeroth-order optimization in the space of precomputed eigenfaces. At every iteration, a batch of random linear combinations (coefficients  $\in N(0,1)$) of eigenfaces is sampled and the best one according to similarity is chosen and added to the reconstructed image which is initialized with 0 at the first iteration. The details of the algorithm can be found in Alg.~\ref{algorithm}.

In \cite{razzhigaev2020blackbox} it was shown that it is very important to control the norm of the embeddings of the generated images and hence use it in the objective function. In this work, we perform the optimization in the space of eigenfaces which, as we suppose, retains the norm of generated images in the correct region. That is why we need just to optimize the similarity without any other information about embeddings or the model itself, which one of the key advantages of our method.

Compared to the Gaussian sampling algorithm our approach does not require the proper initialization.
We would like to highlight that in contrast to previous works \emph{a target face embedding is not used} in the algorithm and it requires only attacked id and a limited number of similarity queries. 

The baseline Gaussian sampling algorithm from \cite{razzhigaev2020blackbox} suffers from the problem of falling into the local minimum producing faces with incorrect appearance. The eigenfaces sampling algorithm also sometimes can have the same behavior, but we found out that using the "multi-start" policy handles this issue (Fig. \ref{compare_multi}). We start the algorithm several times with different random seeds and choose the one with the sharpest optimization trajectory and stop all others. It seems that the speed of convergence at the beginning of the algorithm reflects the final quality of the reconstructed face. Our experiments show that the optimal parameters of the multi-start policy to find the optimal trajectory without a significant increase in the number of queries are the following: 10 starts with 100 iterations in each.

\begin{figure}[t]
\begin{center}
    \includegraphics[width=0.75\linewidth]{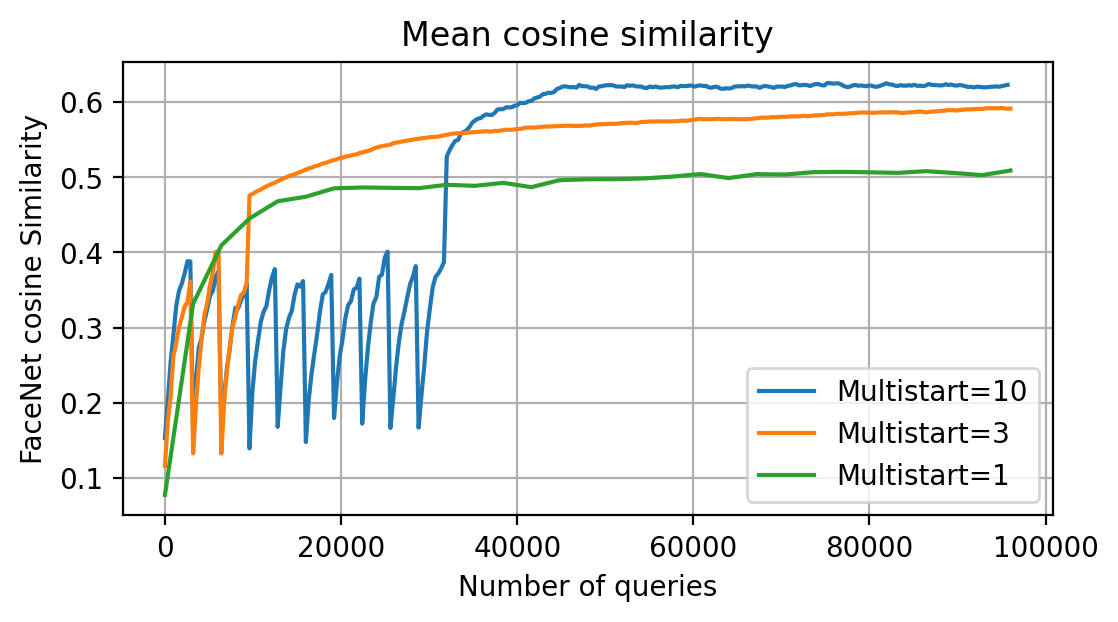}
\end{center}
   \caption{The effect of the multi-start policy. See the text for the details.}
    \label{multi}
\end{figure}

Comparison of dynamics of face recovery procedure with multi-starts and without is shown in Fig.~\ref{multi}. Averaged on 100 random samples from Labeled Faces in the Wild (LFW) dataset~\cite{LFWTech}. The peaks are corresponding to restarts of the algorithm with saving the intermediate reconstruction. After the required number of restarts, the "highest" peak is chosen to continue the optimization and loading the corresponding saved intermediate reconstruction.

\begin{table}[t]
\caption{Comparison with competitors. Average similarity by ArcFace and FaceNet (independent critic) between a reconstructed image and target identity on the LFW dataset. \strut}
\label{results1}
\bigskip
\centering
\begin{tabular}{lccc}
\toprule
Method & ArcFace & FaceNet & \# of queries\\
\midrule
Ours, RGB& $0.97$ & $0.62$& $\textbf{50k}$\\
Ours, gray& $\textbf{0.98}$ & {$\textbf{0.63}$} & $\textbf{50k}$\\
Gaussian blobs, gray& $0.90$ & 0.45 & 300k\\
NBNet, RGB&$0.44$& $0.39$& $3\text{M}$\\
\bottomrule
\end{tabular}
\end{table}

\begin{table}
\parbox{0.48\linewidth}{
\caption{Ablation Study.\\ See text for details.}
\label{ablation}
\bigskip
\centering
\begin{tabular}{lcc}
\toprule
Method & ArcFace & FaceNet \\
\midrule
No R, SL, RGB& $0.89$ & $0.42$ \\
10 R, SL, RGB& $0.96$ & $0.47$ \\
10 R, SR, RGB& 0.97 & 0.53 \\
10 R, GR, RGB& $\textbf{0.98}$ & 0.55 \\
10 R, SR + GR, RGB& $0.97$ & $0.62$\\
10 R, SR + GR, gray& $\textbf{0.98}$ & {$\textbf{0.63}$} \\
\bottomrule
\end{tabular}
}
\hfill
\parbox{0.48\linewidth}{
\caption{LFW verification test results.}
\label{results3}
\bigskip
\centering
\begin{tabular}{lcc}
\toprule[1pt]
Method & ArcFace & FaceNet \\
\midrule[1pt]
Real & $99.71$ & $97.67$ \\
\midrule[0.4pt]
Ours & $\textbf{99.83}$ & $\textbf{94.59}$ \\
NBNet & $91.22$ & $91.01$ \\
DiBiGAN & $99.13$ & $-$ \\
\bottomrule[1pt]
\end{tabular}
}
\end{table}

\section{Experiments}
\subsection{Test methodology}
While validating the effectiveness of considered privacy attacks it is very important to assess the ability of the algorithm to reveal the identity of the person. That is why we follow the previous works: using FaceNet as an independent critic \cite{razzhigaev2020blackbox} and the special LFW verification protocol \cite{duong2020vec2face}.

An independent critic is another neural network in addition to the black-box one for the validation of the face reconstruction results. We attack ArcFace~\cite{deng2019arcface} and use FaceNet~\cite{Schroff_2015} as a model for validation.

\begin{figure}[t]
\begin{center}
   \includegraphics[width=0.9\linewidth]{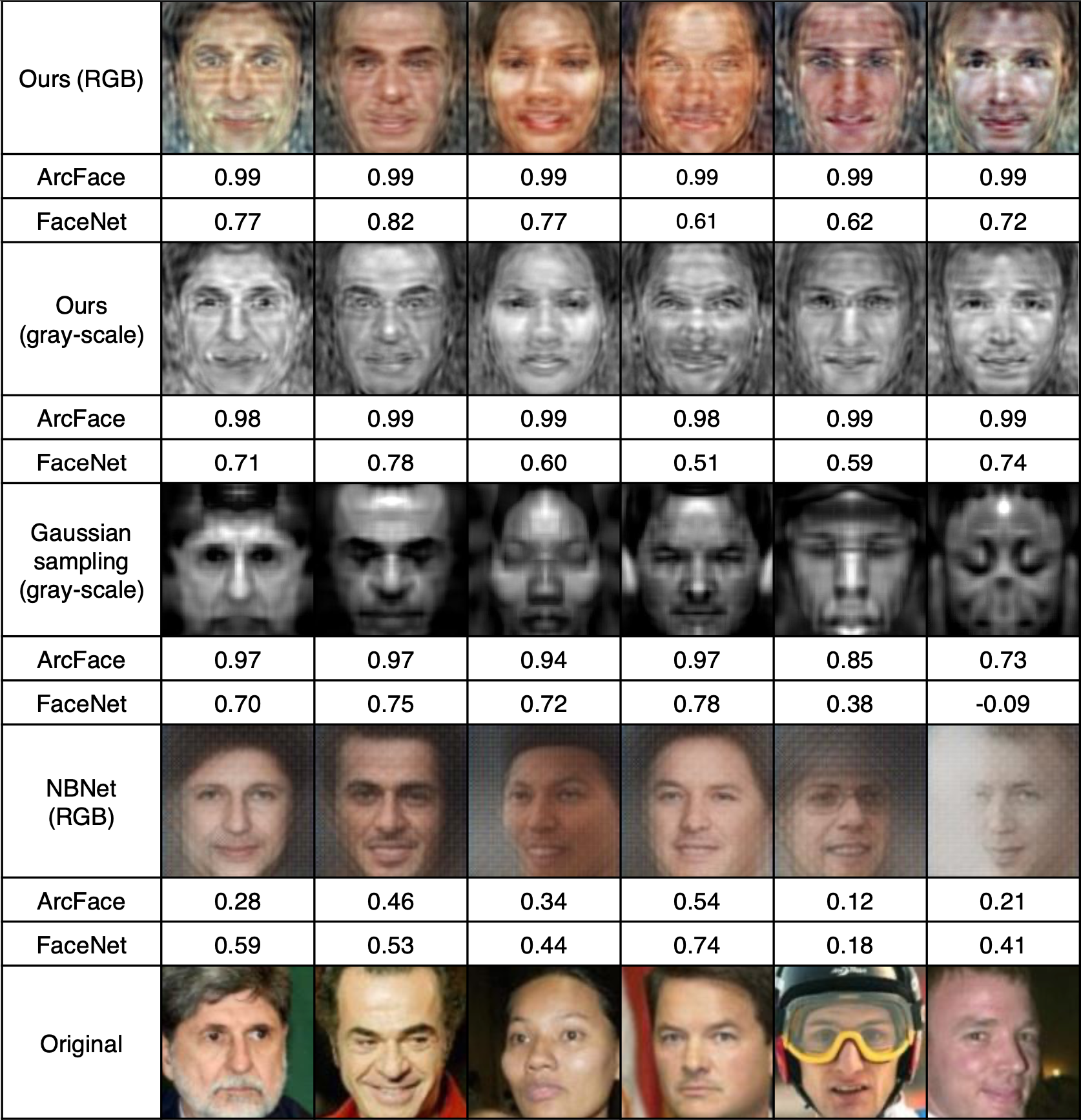}
\end{center}
   \caption{Examples of recovered faces from LFW dataset and corresponding similarities from Arcface and FaceNet.}
\label{compare_all}
\end{figure}

\subsection{Baselines}

We select NBNet \cite{Mai_2019} and Gaussian sampling method \cite{razzhigaev2020blackbox} as our main baselines. We use the original NBNet source code and trained it on the MS1M-ArcFace dataset. The retraining is needed since the original model is trained with the different alignment and worked poorly with photos aligned for ArcFace (by MTCNN \cite{zhang2016joint}). 
In the original paper, the model was trained on the DCGAN output, since there were no sufficient datasets at the time of publication. However, modern datasets are much larger than the number of queries needed for NBNet (e.g., MS1M-ArcFace contains 5.8M images). We follow the original paper training procedure as far as it is possible. The NBNet model is trained with MSE loss at the first stage, then the perceptual loss is added at the second stage. The MSE loss stage took 160K$\times$64 queries, then the loss stopped to decrease. The perceptual loss stage took 100K iterations, as in the original paper. The procedure for Gaussian sampling is taken directly from \cite{razzhigaev2020blackbox}. To compare our method with DiBiGAN\cite{duong2020vec2face} we use the procedure suggested by the author of the original paper. We do not use FaceNet for DiBiGAN evaluation, since the authors did not compare their method with the others on it as well.

\subsection{Results}
We conducted several experiments to validate the performance of our algorithm and for an ablation study.

To test our hypothesis that the first iterations of optimization characterize the total convergence we perform experiments with the different number of multi-starts. We select a random subset of 100 images from LFW and applied our face recovery algorithm while saving the intermediate similarity scores between the reconstructed face and the original one for the black-box model (ArcFace) and the independent critic (FaceNet). The corresponding plots are presented in Fig.~\ref{multi}. Increasing the number of multi-starts increases the slope of convergence and the final score. Without restarts optimization sometimes gets stuck in the local minima corresponding to the wrong identity Fig.~\ref{compare_multi}. We have noticed that if the reconstructed identity is correct -- the optimization convergence is faster (even at the very first iterations) which led us to exploit restarting algorithm multiple times and to choose the optimal trajectory avoiding local minima. Our experiments show that 10 restarts with 100 iterations at each one are enough to avoid almost all local minima.

For the ablation study, we used 1055 images from LFW (subset A).
We computed average similarity by ArcFace and FaceNet (independent critic) between a reconstructed image and target identity.
As it is seen in Table~\ref{ablation}: multi-starts (R), symmetry regularization (SR), and generative regularization (GR) give improvement to the final performance in comparison with standard MSE loss (SL) of the autoencoder and recovery without restarts.

In Table~\ref{results1} you can find the final similarity scores between original images and reconstructed faces averaged on the LFW dataset. Our algorithm outperforms the baselines not only in the number of queries to the black-box model but in the final similarity scores (for both the black-box model and the independent critic) which demonstrates that it is possible to reveal the identity of a person from a face recognition system by just querying similarity scores. In Table~\ref{results3} there are verification test scores on LFW for our method and competitors. As one can notice, for our algorithm the ArcFace verification score on the reconstructed faces is even higher than for real faces. We suggest that it is an inference-time preprocessing effect making images "cleaner", removing occlusions, and simultaneously retaining a pretty high similarity score (0.98) with the initial images.

Examples of recovered faces for different face reconstruction methods can be found in Fig. \ref{compare_all}.

\section{Conclusion}
We are the first, to the best of our knowledge, to demonstrate the possibility to recover recognizable faces from open-set face recognition systems querying only similarity scores without any knowledge about face embeddings or model architecture according to a fully black-box setup. While our method works in the toughest conditions, it requires a smaller number of queries to the attacked model and outperforms previous methods in terms of mean similarity scores between recovered and original faces as well as in terms of LFW verification rate.

\bibliography{biblio}

\end{document}